# Topic Identification in Discourse

**Kuang-hua Chen**

Department of Computer Science and Information Engineering

National Taiwan University

Taipei, Taiwan, R.O.C.

khchen@nlg.csie.ntu.edu.tw

**Abstract**

This paper proposes a corpus-based language model for topic identification. We analyze the association of noun-noun and noun-verb pairs in LOB Corpus. The word association norms are based on three factors: 1) word importance, 2) pair co-occurrence, and 3) distance. They are trained on the paragraph and sentence levels for noun-noun and noun-verb pairs, respectively. Under the topic coherence postulation, the nouns that have the strongest connectivities with the other nouns and verbs in the discourse form the preferred topic set. The collocational semantics then is used to identify the topics from paragraphs and to discuss the topic shift phenomenon among paragraphs.

## 1 Introduction

Although only speakers and writers instead of texts have topics (Brown and Yule, 1983, p. 68), natural language researchers always want to identify a topic or a set of possible topics from a discourse for further applications, such as anaphora resolution, information retrieval and so on. This paper adopts a corpus-based approach to process discourse information. We postulate that:

(1) *Topic is coherent and has strong relationships with the events in the discourse.*

Now, consider the following example quoted from the Lancaster-Oslo/Bergen (LOB) Corpus (Johansson, 1986). The topics in this example are "problem" and "dislocation". The two words are more strongly related to the verbs ("explain", "fell", "placing" and "suppose") and nouns ("theories", "explanations", "roll", "codex", "disorder", "order", "disturbance" and "upheaval").

> There is a whole group of theories which attempt to explain the problems of the Fourth Gospel by explanations based on assumed textual dislocations. The present state of the Gospel is the result of an accident-prone history. The original was written on a roll, or codex, which fell into disorder or was accidentally damaged. An editor, who was not the author, made what he could of the chaos by placing the fragments, or sheets, or pages, in order. Most of those who expound a theory of textual dislocation take it for granted that the Gospel was written entirely by one author before the disturbance took place but a few leave it open to suppose that the original book had been revised even before the upheaval.

We also postulate that

(2) *Noun-verb is a predicate-argument relationship on the sentence level and noun-noun relationship is associated on discourse level.*

The postulation (2) could be also observed from the above example. These relationships may be represented implicitly by collocational semantics. Collocation has been applied successfully to many possible applications (Church *et al.* , 1989), e.g, lexicography (Church and Hanks, 1990), information retrieval (Salton, 1986a), text input (Yamashina and Obashi, 1988), etc. This paper will touch on its feasibility in topic identification.

This paper is organized as follows. Section 2 presents a corpus-based language model and discuss how to train this model. Section 3 touches on topic identification in discourse. Section 4 shows a series of experiments based on the proposed model and discusses the results. Section 5 gives short remarks.

## 2 A Language Model

Brown and Yue (1983) pointed out there are two kinds of topics: one is sentence topic and the other is discourse topic. The discourse topic is usually the form of topic sentence. We postulate, further, that the noun in the topic sentence play important roles in the whole discourse. Thus nouns play the core part in the underlying language model. The associations of a noun with other nouns and verbs are supporting factors for it to be a topic.

The importance of a specific verb or noun is defined by Inverse Document Frequency (*IDF*) (Salton, 1986b):

$$IDF(W) = \log((P - O(W))/O(W)) + c \qquad (1)$$

where *P* is the number of documents in LOB Corpus, i.e. 500, *O(W)* is the number of documents with word *W*, and *c* is a threshold value. LOB Corpus is a million-word collection of present-day British English texts. It contains 500 texts of approximately 2,000 words distributed over 15 text categories (Johansson, 1986). These categories include reportage, editorial, reviews, religion, skills, trades, popular lore, belles lettres, biography, essays, learned and scientific writings, fictions, humour, adventure and western fiction, love story, *etc*. That is to say, LOB Corpus is a balanced corpus. The tag set of LOB Corpus is based on the philosophy of that of Brown Corpus (Francis and Kucera, 1979), but some modifications are made. This is to achieve greater delicacy, while preserving comparability with the Brown Corpus.

Those words that appear more than one half of the documents in LOB Corpus have negative log((P-O(W))/O(W)) shown below.

Noun: time(-3.68)  way(-1.92)  year(-1.71)
      man(-1.47)  day(-1.12)  part(-0.76)
      people(-0.75)  thing(-0.73)  hand(-0.54)
      life(-0.51)  fact(-0.40)  place(-0.40)
      work(-0.35)  end(-0.12)  case(-0.09)
      point(-0.05)

Verb: make(-5.01)  take(-3.56)  give(-2.95)
      come(-2.45)  find(-2.30)  see(-2.26)
      know(-2.20)  say(-2.18)  go(-2.11)
      seem(-1.30)  show(-1.20)  think(-1.18)
      use(-1.07)  get(-1.06)  become(-0.95)
      bring(-0.73)  put(-0.68)  leave(-0.62)
      look(-0.48)  call(-0.43)  tell(-0.41)
      keep(-0.32)  hold(-0.18)  ask(-0.23)
      begin(-0.08)

The threshold values for nouns and verbs are set to 0.77 and 2.46 respectively. The two values are used to screen out the *unimportant* words, whose *IDF* values are negative. That is, their *IDF* values are reset to zero. The strength of one occurrence of a verb-noun pair or a noun-noun pair is computed by the importance of the words and their distances:

$$SNV(N_i, V_j) = IDF(N_i) \cdot IDF(V_j) / D(N_i, V_j) \qquad (2)$$

$$SNN(N_i, N_k) = IDF(N_i) \cdot IDF(N_k) / D(N_i, N_k) \qquad (3)$$

where *SNV* denotes the strength of a noun-verb pair, *SNN* the strength of a noun-noun pair, and *D(X,Y)* represents the distance between *X* and *Y*. When *i* equals to *k*, the *SNN(N_i,N_k)* is set to zero. Including the distance factor is motivated by the fact that the related events are usually located in the same texthood. This is the spatial locality of events in a discourse.

The distance is measured by the difference between cardinal numbers of two words. We assign a cardinal number to each verb and noun in sentences. The cardinal numbers are kept continuous across sentences in the same paragraph. For example,

With so many problems$_1$ to solve$_2$, it would be a great help$_3$ to select$_4$ some one problem$_5$ which might be the key$_6$ to all the others, and begin$_7$ there. If there is any such key-problem$_8$, then it is undoubtedly the problem$_9$ of the unity$_{10}$ of the Gospel$_{11}$. There are three views$_{12}$ of the Fourth Gospel$_{13}$ which have been held$_{14}$.

Therefore, the cardinal number of *problems*, *C(problems)*, equals to 1 and *C(held)* is 14. The distance can be defined to be

$$D(X,Y) = abs(C(X)-C(Y)) \qquad (4)$$

The association norms of verb-noun and noun-noun pairs are summation of the strengths of all their occurrences in the corpus:

$$ANV(N_i, V_j) = \sum SNV(N_i, V_j) \qquad (5)$$

$$ANN(N_i, N_k) = \sum SNN(N_i, N_k) \qquad (6)$$

where *ANV* denotes the association norm of a noun-verb pair, and *ANN* the association norm of a noun-noun pair. The less frequent word has a higher *IDF* value so that the strength *SNV* and *SNN* of one occurrence may be larger. However, the number of terms to be summed is smaller. Thus, the formulae *IDF* and *ANV* (*ANN*) are complementary. LOB Corpus of approximately one million words is used to train the basic association norms. They are based on different levels: the paragraph and sentence levels for noun-noun and noun-verb pairs respectively. Table 1 shows the statistics of the training corpus. The words with tags NC, NNU and NNUS and Ditto tags are not considered. Here NC means cited words, and NNU (NNUS) denotes abbreviated (plural) unit of measurement unmarked for number. Ditto tags are those words whose senses in combination differ from the role of the same words in other context. For example, "as to", "each other", and "so as to" (Johansson, 1986).

**Table 1. Statistics for LOB Corpus**

|  | number |
|---|---|
| Document | 500 |
| Paragraph | 18678 |
| Sentences | 54297 |
| Nouns | 23399 |
| Verbs | 4358 |
| N-N Pairs | 3476842 |
| V-N Pairs | 422945 |

Under the topic coherence postulation in a paragraph, we compute the connectivities of the nouns in each sentence with the verbs and nouns in the paragraph. For example, 439 verbs in LOB Corpus have relationships with the word "problem" in different degrees. Some of them are listed below in descending order by the strength.

solve(225.21), face(84.64), ..., specify(16.55), ..., explain(6.47), ..., fall(2.52), ..., suppose(1.67), ...

For the example in Section 1, the word "problem" and "dislocation" are coherent with the verbs and nouns in the discourse. The nouns with the strongest connectivity form the preferred topic set. Consider the interference effects. The constituents far apart have less relationship. Distance $D(X,Y)$ is used to measure such effects. Assume there are $m$ nouns and $n$ verbs in a paragraph. The connective strength of a noun $N_i$ ($1 \leq i \leq m$) is defined to be:

$$CSNN(N_i) = \sum_k (ANN(N_i, N_k) / D(N_i, N_k)) \quad (7)$$

$$CSNV(N_i) = \sum_k (ANN(N_i, V_k) / D(N_i, V_k)) \quad (8)$$

$$CS(N_i) = (PN \cdot CSNN(N_i) + PV \cdot CSNV(N_i)) / c \quad (9)$$

where $CS$ denotes the connective strength, and $PN$ and $PV$ are parameters for $CSNN$ and $CSNV$ and $PN+PV=1$.

The determination of $PN$ and $PV$ is via deleted interpolation (Jelinek, 1985). Using equation $PN + PV = 1$ and equation 9, we could derive $PN$ and $PV$ as equation 10 and equation 11 show.

$$PN = \frac{CS - CSNV}{CSNN - CSNV} \quad (10)$$

$$PV = \frac{CS - CSNN}{CSNV - CSNN} \quad (11)$$

LOB corpus are separated into two parts whose volume ratio is 3:1. Both $PN$ and $PV$ are initialized to 0.5 and then are trained by using the 3/4 corpus. After the first set of parameters is generated, the remain 1/4 LOB corpus is run until $PN$ and $PV$ converge using equations 9, 10 and 11. Finally, the parameters, $PN$ and $PV$, converge to 0.675844 and 0.324156 respectively.

## 3 Topic Identification in a Paragraph

The test data are selected from the first text of the files LOBT-D1, LOBT-F1, LOBT-G1, LOBT-H1, LOBT-K1, LOBT-M1 and LOBT-N1 of horizontal version of LOB Tagged Corpus for inside test (hereafter, we will use D01, F01, G01, H01, K01, M01, and N01 to represent these texts respectively). Category D denotes religion, Category F denotes popular lore, Category G denotes belles lettres, biography and essays, Category H denotes Miscellaneous texts, Category K denotes general fiction, Category M denotes science fiction, and Category N denotes adventure and western fiction. Each paragraph has predetermined topics (called *assumed* topics) which are determined by a linguist. Because a noun with basic form $N$ may appear more than once in the paragraph, say $k$ times, its strength is normalized by the following recursive formula:

$$NCS(N_{o(1)}) = CS(N_{o(1)}) \quad (12)$$

$$NCS(N_{o(i)}) = NCS(N_{o(i-1)}) + (1 - NCS(N_{o(i-1)})) \cdot CS(N_{o(i)}) \quad (13)$$

where $NCS$ represents the net connective strength, $o(k)$ denotes the cardinal number of the $k$'th occurrence of the same $N$ such that $C(N_{o(1)}) < C(N_{o(2)}) < C(N_{o(3)}) < ... < C(N_{o(k-1)}) < C(N_{o(k)})$.

The possible topic $N^*$ has the high probability $NCS(N^*)$. Here, a topic set whose members are the first 20% of the candidates is formed. The performance can be measured as the Table 2 shows.

## 4 The Preliminary Experimental Results

According to the language model mentioned in Section 2, we build the $ANN$ and $ANV$ values for each noun-noun pair and noun-verb pair. Then, we apply recursive formula of $NCS$ shown in equations 12 and 13 to identifying the topic set for test texts. Table 3 shows experimental results. Symbols μ and σ denotes mean and standard deviation. (+) denotes correct number, (-) denotes error number and (?) denotes undecidable number in topic identification. The undecidable case is that the assumed topic is a pronoun. Figure 1 shows correct rate, error rate, and undecidable rate.

Row (1) in Table 3 shows the difficulty in finding topics from many candidates. In general, there are more than 20 candidates in a paragraph. It is impossible to select topics at random. Row (2) gives the rank of assumed topic. The assumed topics are assigned by a linguist. Comparing row (1) and row (2), the average number of candidates are much larger than the rank of assumed topic. Since it is impossible to randomly select candidates as topics, we know topic identification is valuable.

Rows (3), (4) and (5) list the frequencies of candidates, assumed topics and computed topic. The results intensify the viewpoint that the repeated words make persons impressive, and these words are likely to be topics. Our topic identification algorithm demonstrates the similar behavior (see rows (4) and (5)). The average frequencies of assumed topics and computed topics are close and both of them are larger than average frequency of candidates. Figure 2 clearly demonstrates this point. Row (6) reflects an interesting phenomenon. The topic shifted by authors from paragraph to paragraph is demonstrated through comparison of data shown in this row and row (2). The rank value of previous topics do obviously increase. Recall that large rank value denotes low rank.

### Table 2. Metrics for Performance

| | | |
|---|---|---|
| 1 | average # of candidates | Σ # of nouns in basic form in paragraph i / # of paragraphs |
| 2 | average rank of assumed topic | Σ rank of assumed topic in paragraph i / # of paragraphs |
| 3 | frequency of candidates | Σ # of nouns / Σ # of nouns in basic form in paragraph i |
| 4 | frequency of assumed topic | Σ occurrences of assumed topic / # of paragraphs |
| 5 | frequency of computed topic | Σ occurrences of computed topic / # of paragraphs |
| 6 | average rank of topic in previous paragraph | Σ rank of topic in previous paragraph / (# of paragraph - 1) |

### Table 3. Experimental Results

| (μ, σ) | D01 | F01 | G01 | H01 | K01 | M01 | N01 |
|---|---|---|---|---|---|---|---|
| (1) | (21.59, 9.96) | (10.57, 18.42) | (62.43, 18.42) | (19.77, 8.39) | (31.71, 23.80) | (15.22, 6.44) | (12.21, 6.73) |
| (2) | (4.56, 5.98) | (5.25, 5.51) | (7.29, 10.35) | (4.55, 4.13) | (7.08, 16.02) | (2.61, 2.11) | (3.68, 3.87) |
| (3) | (1.32, 0.88) | (1.39, 0.89) | (1.21, 0.56) | (1.33, 0.82) | (1.11, 0.39) | (1.11, 0.32) | (1.06, 0.25) |
| (4) | (2.61, 1.60) | (1.27, 1.21) | (2.57, 1.18) | (2.46, 1.62) | (1.77, 1.05) | (1.50, 0.69) | (1.28, 0.60) |
| (5) | (3.33, 1.97) | (2.39, 1.84) | (3.43, 1.40) | (2.91, 1.56) | (1.86, 0.99) | (1.48, 0.50) | (1.29, 0.52) |
| (6) | (6.29, 7.84) | (5.48, 5.09) | (19.67, 16.64) | (5.71, 6.06) | (17.23, 18.51) | (7.92, 6.28) | (9.36, 6.62) |
| (+) | 12 | 13 | 6 | 12 | 9 | 13 | 15 |
| (-) | 6 | 15 | 1 | 10 | 4 | 5 | 10 |
| (?) | 0 | 0 | 0 | 0 | 1 | 9 | 9 |

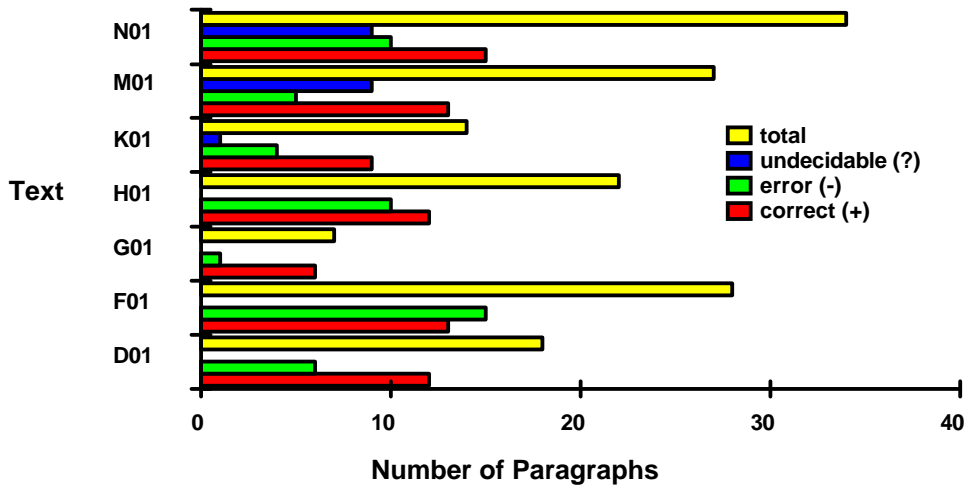

Figure 1. The Results of Topic Identification

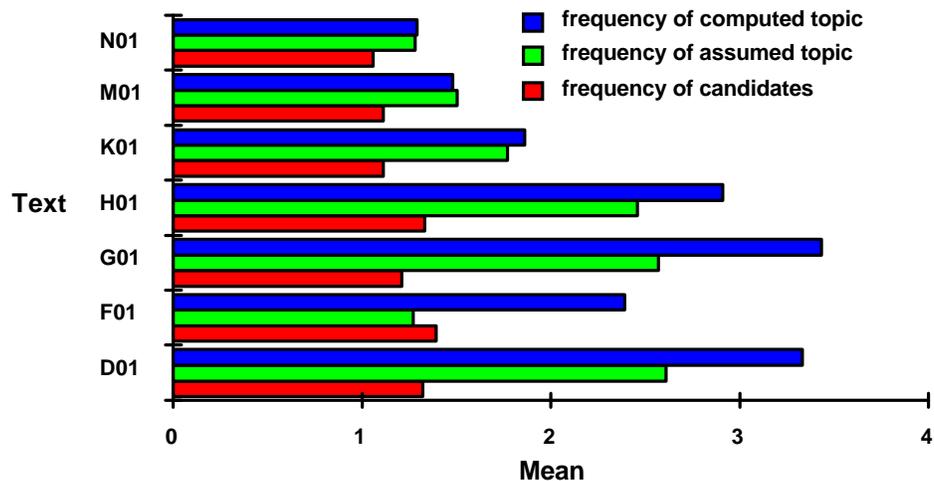

Figure 2. Comparison of Frequency

## 5 Concluding Remarks

Discourse analysis is a very difficult problem in natural language processing. This paper proposes a corpus-based language model to tackle topic identification. The word association norms of noun-noun pairs and noun-verb pairs which model the meanings of texts are based on three factors: 1) word importance, 2) pair occurrence, and 3) distance. The nouns that have the stronger connectivities with other nouns and verbs in a discourse could form a preferred topic set. Inside test of this proposed algorithm shows 61.07% correct rate (80 of 131 paragraphs).

Besides topic identification, the algorithm could detect topic shift phenomenon. The meaning transition from paragraph to paragraph could be detected by the following way. The connective strengths of the topics in the previous paragraph with the nouns and the verbs in the current paragraph are computed, and compared with the topics in the current paragraph. As our experiments show, the previous topics have the tendency to decrease their strengths in the current paragraph.

## Acknowledgment

We are thankful to Yu-Fang Wang and Yue-Shi Lee for their help in this work.